\documentclass{article} 
\usepackage{ijcai20}

\usepackage{times}
\usepackage{soul}
\usepackage{url}
\usepackage{multirow}
\usepackage[hidelinks]{hyperref}
\usepackage[utf8]{inputenc}
\usepackage[small]{caption}
\usepackage{graphicx}
\usepackage{subcaption}
\usepackage{amsmath}
\usepackage{amsthm}
\usepackage{booktabs}
\usepackage{algorithm}
\usepackage{algorithmic}
\usepackage{xcolor}
\usepackage{hyperref}
\usepackage{url}
\usepackage{graphicx}
\usepackage{amsmath}
\usepackage{bbm}

\newcommand{\mb}{\mathbf}
\newcommand{\bs}{\boldsymbol}
\newcommand{\mc}{\mathcal}

\newtheorem{definition}{\textsc{Definition}}

\newcommand{\our}{\textsc{DifNet}}
\newcommand{\gdu}{\textsc{GDU}}
\newcommand{\loopy}{\textsc{LoopyNet}}

\newcommand{\gcn}{\textsc{GCN}}
\newcommand{\gat}{\textsc{GAT}}

\title{Get Rid of Suspended Animation Problem: Deep Diffusive Neural Network on Graph Semi-Supervised Classification}

\author{Jiawei~Zhang\
\affiliations
IFM Lab, Florida State University, Tallahassee, FL, USA\\
 \emails
  jiawei@ifmlab.org}

\begin{document}

\maketitle

\begin{abstract}

Existing graph neural networks may suffer from the ``suspended animation problem'' when the model architecture goes deep. Meanwhile, for some graph learning scenarios, e.g., nodes with text/image attributes or graphs with long-distance node correlations, deep graph neural networks will be necessary for effective graph representation learning. In this paper, we propose a new graph neural network, namely {\our} (Graph Diffusive Neural Network), for graph representation learning and node classification. {\our} utilizes both neural gates and graph residual learning for node hidden state modeling, and includes an attention mechanism for node neighborhood information diffusion. Extensive experiments will be done in this paper to compare {\our} against several state-of-the-art graph neural network models. The experimental results can illustrate both the learning performance advantages and effectiveness of {\our}, especially in addressing the ``suspended animation problem''.

\end{abstract}
\section{Introduction}\label{sec:introduction}

Graph neural networks (GNNs) \cite{Wu_A_19,Zhang_Graph_19} have been one of the most important research topic in recent years, and many different types of GNN models have been proposed already \cite{Kipf_Semi_CORR_16}. Generally, via the aggregations of information from the neighbors \cite{Hammond_Wavelets_11,Defferrard_Convolutional_16}, GNNs can learn the representations of nodes in graph data effectively. The existing GNNs, e.g., {\gcn} \cite{Kipf_Semi_CORR_16}, {\gat} \cite{Velickovic_Graph_ICLR_18} and their derived variants \cite{Klicpera_Personalized_18,Li_Combinatorial_18,Gao_GraphNAS_19}, are mostly based on the approximated graph convolutional operator \cite{Hammond_Wavelets_11,Defferrard_Convolutional_16}. As introduced in \cite{Zhang_GResNet_19}, such an approximated aggregation operator actually reduces the neighboring information integration as an one-hop random walk layer followed by a fully-connected layer. Generally, in many learning scenarios, e.g., nodes with text/image attributes or graphs with close correlations among nodes that are far away, a deep GNN architecture will be necessary to either achieve the high-order latent attribute representations or integrate information from those long-distance nodes for graph representation learning.

\begin{figure}
    \centering
    \begin{subfigure}[b]{.23\textwidth}
    	\includegraphics[width=\linewidth]{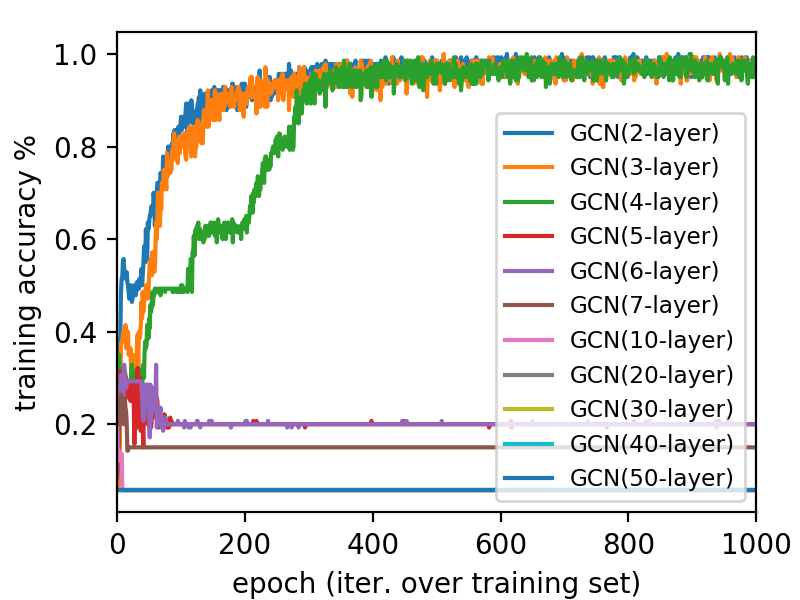}
    	\caption{Training Accuracy}\label{fig:gcn_acc_train}
    \end{subfigure}%
    \hfill
    \begin{subfigure}[b]{.23\textwidth}
    	\includegraphics[width=\linewidth]{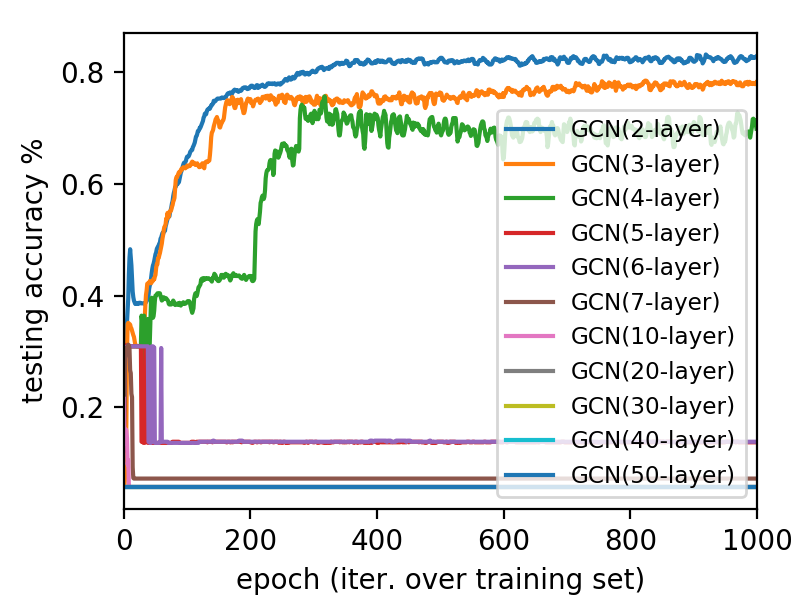}
    	\caption{Testing Accuracy}\label{fig:gcn_acc_test}
    \end{subfigure}%
    \caption{The learning performance of {\gcn} (bias disabled) with 2-layer, $\dotsc$, 7-layer, and 10-layer, $\cdots$, 50-layer on the Cora dataset. The x axis denotes the iterations over the whole training set. The y axes denote the training and testing accuracy, respectively.}\label{fig:gcn_acc_analysis}
\end{figure}

Also as pointed out in \cite{Zhang_GResNet_19}, most of the existing GNN models based on approximated graph convolutional operator will suffer from the ``suspended animation problem'' when the model architecture goes deep. Especially, when the GNNs' model depth reaches certain limit (namely, the GNNs' suspended animation limit), the model will not respond to the training data anymore and become not learnable. As illustrate in Figure~\ref{fig:gcn_acc_analysis}, we show a case study of the vanilla {\gcn} model (with bias term disabled) on the Cora dataset (a well-known benchmark graph dataset). According to the results, the model can achieve the best performance with 2 layers, and its performance gets degraded as the depth increases and further get choked as the depth reaches $5$ and more. Therefore, 5-layer is also called the ``suspended animation limit'' of {\gcn}. Similar phenomena have been observed for the {\gcn} model (with bias term enabled), {\gat} and their other derived models as well.

In this paper, we will introduce a novel graph neural network model, namely {\our} (Graph Diffusive Neural Net), for graph representation learning. {\our} is not based on the approximated graph convolutional operator and can work perfectly with deep model architectures. To be more specific, {\our} introduces a new neuron unit, i.e., {\gdu} (gated diffusive unit), for modeling and updating the graph node hidden states at each layer. Equipped with both neural gates and graph residual learning techniques, {\gdu} can help address the aforementioned ``suspended animation problem'' effectively. Furthermore, {\our} also involves the attention mechanism for the neighboring node representation diffusion prior to feeding them into {\gdu}. All the variables involved in {\gdu} can be learned automatically with the backpropagation algorithm efficiently.

\begin{figure}
    \centering
    \begin{subfigure}[b]{.23\textwidth}
    	\includegraphics[width=\linewidth]{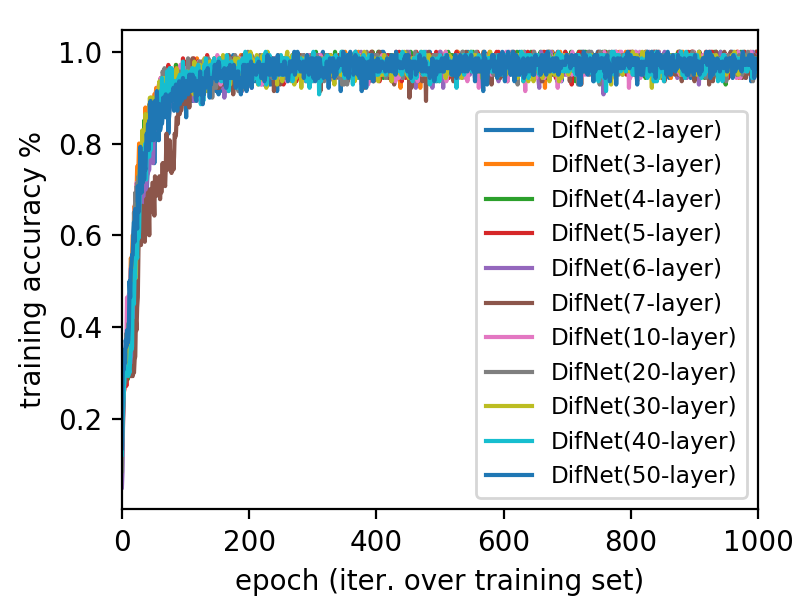}
    	\caption{Training Accuracy}\label{fig:difnet_acc_train}
    \end{subfigure}%
    \hfill
    \begin{subfigure}[b]{.23\textwidth}
    	\includegraphics[width=\linewidth]{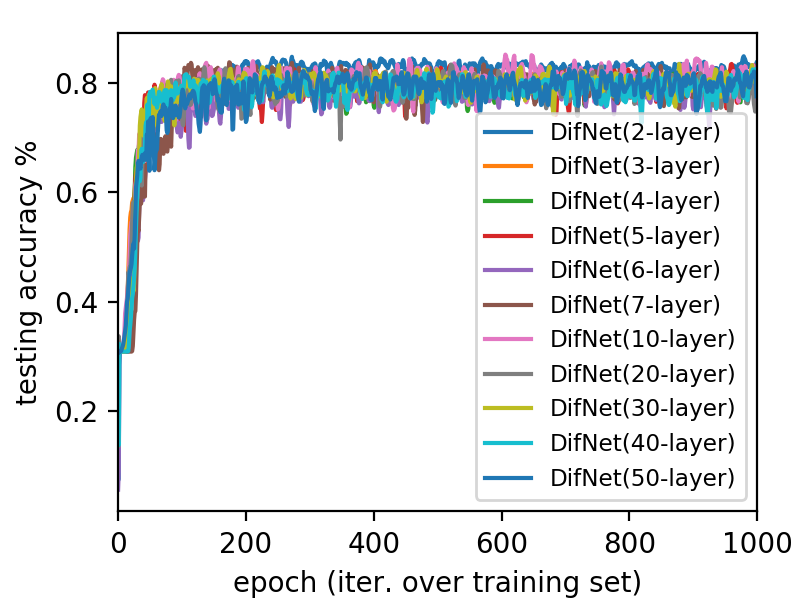}
    	\caption{Testing Accuracy}\label{fig:difnet_acc_test}
    \end{subfigure}%
    \caption{{The learning performance of {\our} with 2-layer, 3-layer, $\dotsc$, 50-layer on the Cora dataset.}}\label{fig:difnet_acc_analysis}
\end{figure}

To demonstrate the effectiveness of {\our}, in Figure~\ref{fig:difnet_acc_analysis}, we also illustrate the learning performance of {\our} on the Cora dataset (all the settings are identical to those of Figure~\ref{fig:gcn_acc_analysis}). From the plots, we observe that {\our} proposed in this paper can converge very fast and also outperforms {\gcn} for the testing results. What's more, {\our} generalizes very well to the deep architectures, and its learning performance can even get improved as the model depth increases. For instance, among all these layers, the highest testing accuracy score is achieved by {\our} (10-layer). To be more specific, the layers for {\our} denotes the number of {\gdu} based layers. More detailed information and experimental results about {\our} will be provided in the following sections of this paper.

We summarize the contributions of this paper as follows:
\begin{itemize}
\item We introduce a novel graph neural network model {\our} in this paper, which can help address the ``suspended animation problem'' effectively on graph representation learning.
\item We propose a new neuron model, i.e., {\gdu}, which involves both neural gates and residual learning for node representation modeling and updating.
\item {\our} introduces an attention mechanism for the node representation diffusion and integration, which can work both effectively and efficiently. 
\item We test the effectiveness of {\our} on several commonly used graph benchmark datasets and compare against both classic and state-of-the-art GNN models.
\end{itemize}

The remaining parts of this paper will be organized as follows. We will introduce the related work in Section~\ref{sec:related_work}. Definitions of important terminologies and formulation of the studied problem will be provided in Section~\ref{sec:formulate}. Detailed information about the {\our} model will be introduced in Section~\ref{sec:method}. The effectiveness of {\our} will be tested in Section~\ref{sec:experiment}. Finally, we will conclude this paper in Section~\ref{sec:conclusion}.

\section{Related Work}\label{sec:related_work}

Several research topics are closely correlated with this paper, which include \textit{graph neural network}, \textit{GNN learning problems}, \textit{neural gates} and \textit{residual learning}, respectiely.

\noindent \textbf{Graph Neural Network}: Due to the inter-connected structure of graph data, traditional deep models (assuming the training instances are i.i.d.) cannot be directly applied to the graph data. In recent years, graph neural networks \cite{MBMRSB16,AT16,MBBV15,Kipf_Semi_CORR_16,SGTHM09,ZCZYLS18,NAK16} have become one of the most popular research topic for effective graph representation learning. GCN proposed in \cite{Kipf_Semi_CORR_16} feeds the generalized spectral features into the convolutional layer for representation learning. Similar to {\gcn}, deep loopy graph neural network \cite{loopynet} proposes to update the node states in a synchronous manner, and it introduces a spanning tree based learning algorithm for training the model. {\loopy} accepts nodes' raw features into each layer of the model, and it doesn't suffer from the suspended animation problem according to \cite{Zhang_GResNet_19}. GAT \cite{Velickovic_Graph_ICLR_18} leverages masked self-attentional layers to address the shortcomings of GCN. In recent years, we have also observed several derived variants of GCN and GAT models, e.g., \cite{Klicpera_Personalized_18,Li_Combinatorial_18,Gao_GraphNAS_19}. Due to the limited space, we can only name a few number of the representative graph neural network here. The readers are also suggested to refer to page\footnote{https://paperswithcode.com/task/node-classification}, which provides a summary of the latest graph neural network research papers with code on the node classification problem. 

\noindent \textbf{GNNs' Performance Problems}: For the existing graph neural networks, several performance problems have been observed. In \cite{Zhang_GResNet_19}, the unique ``suspended animation problem'' with graph neural networks is identified and formally defined. A theoretic analysis about the causes of such a problem is also provided in \cite{Zhang_GResNet_19}, which also introduce the graph residual learning as the potential solution. Several other problems have also been observed in learning deep graph neural networks, e.g., the ``over-smoothing problem'' \cite{Li_Deeper_18}. To resolve such a problem, many research efforts have been witnessed \cite{Li_Can_19,Merve_An_19,Sun_AdaGCNAG_19,Huang_Inductive_19}. To enable deep GCNs, \cite{Li_Can_19} proposes to adopt residual/dense connections and dilated convolutions into the GCN architecture. \cite{Merve_An_19} tries to provide an explanation about the cause why GCN cannot benefit from deep architectures. In \cite{Sun_AdaGCNAG_19}, the authors propose a novel RNN-like deep graph neural network architecture by incorporating AdaBoost into the computation of network. Similar methods is also observed in \cite{Huang_Inductive_19}, which also extends the recurrent network for deep graph representation learning.

\noindent \textbf{Neural Gates and Residual Learning}: Besides these two main topics, this paper is also closely correlated with \textit{neural gates learning} \cite{Hochreiter_Long_97,Chung_Empirical_14} and \textit{residual learning} \cite{Kumar_Highway_15,He_Deep_15,Zhang_GResNet_19}, respectively. Neural gate learning initially proposed in LSTM \cite{Hochreiter_Long_97} proposes a way to handle the gradient vanishing/exploding problem with deep model learning. Later on, such concepts have been widely used in deep model learning \cite{Chung_Empirical_14,Gao2016DeepGR}. Residual network \cite{He_Deep_15} has been demonstrated to be effective for deep CNN learning, which extends the high-way network \cite{Kumar_Highway_15} to a more general case. In \cite{Zhang_GResNet_19}, the authors also observe that conventional naive residual learning cannot work for graph neural networks any more, and they introduce several new graph residual terms for representation learning instead.  
\section{Problem Formulation}\label{sec:formulate}

Here, we will provide the formulation of the problem studied in this paper. Prior to that, we will also introduce the notations and terminologies to be used in the problem formulation

\subsection{Notations}\label{subsec:notation}

In the sequel of this paper, we will use the lower case letters (e.g., $x$) to represent scalars, lower case bold letters (e.g., $\mb{x}$) to denote column vectors, bold-face upper case letters (e.g., $\mb{X}$) to denote matrices, and upper case calligraphic letters (e.g., $\mathcal{X}$) to denote sets or high-order tensors. Given a matrix $\mb{X}$, we denote $\mb{X}(i,:)$ and $\mb{X}(:,j)$ as its $i_{th}$ row and $j_{th}$ column, respectively. The ($i_{th}$, $j_{th}$) entry of matrix $\mb{X}$ can be denoted as either $\mb{X}(i,j)$. We use $\mb{X}^\top$ and $\mb{x}^\top$ to represent the transpose of matrix $\mb{X}$ and vector $\mb{x}$. For vector $\mb{x}$, we represent its $L_p$-norm as $\left\| \mb{x} \right\|_p = (\sum_i |\mb{x}(i)|^p)^{\frac{1}{p}}$. The Frobenius-norm of matrix $\mb{X}$ is represented as $\left\| \mb{X} \right\|_F = (\sum_{i,j} |\mb{X}(i,j)|^2)^{\frac{1}{2}}$. The element-wise product of vectors $\mb{x}$ and $\mb{y}$ of the same dimension is represented as $\mb{x} \otimes \mb{y}$, whose concatenation is represented as $\mb{x} \sqcup \mb{y}$.

\subsection{Terminology Definition}

In this paper, we focus on studying the node representation learning problem in graph data.

\begin{definition}
(Graph): Formally, we can denote the graph data studied in this paper as $G = (\mc{V}, \mc{E}, w, x, y)$, where $\mc{V}$ and $\mc{E}$ denote the sets of nodes and links in the graph. Mapping $w: \mc{E} \to \mathbbm{R}$ projects the links to their weights; whereas mappings $x: \mc{V} \to \mc{X}$ and $y: \mc{V} \to \mc{Y}$ project the nodes to their raw features and labels. Here, $\mc{X}$ and $\mc{Y}$ denote the raw feature space and label space, respectively.
\end{definition}

According to the above definition, for presentation simplicity, given a node $v_i \in \mc{V}$, we can also denote its raw feature vector and label vector as $\mb{x}_i = x(v_i) \in \mathbbm{R}^{d_x}$ and $\mb{y}_i = y(v_i)  \in \mathbbm{R}^{d_y}$, respectively. Meanwhile, given a link $(v_i, v_j) \in \mc{E}$ between nodes $v_i$ and $v_j$, we can also denote its corresponding link weight as $w(v_i, v_j)$ (or $w_{i,j}$ for simplicity). In the case that studied graph is unweighted, we have $w_{i,j} = 1, \forall (v_i, v_j) \in \mc{E}$ and $w_{i,j} = 0, \forall (v_i, v_j) \in \mc{V} \times \mc{V} \setminus \mc{E}$.

\begin{definition}
(Neighbor Set): Given the input graph $G$, for any node $v_i \in \mc{V}$ in the graph, we can denote its neighbor set as the group of nodes adjacent to $v_i$ based on the links in the graph, i.e., $\Gamma(v_i) = \{v_j | v_j \in \mc{V} \land (v_i, v_j) \in \mc{E} \}$.
\end{definition}

In the case when the studied graph is directed, the neighborhood set can be furthermore divided into the the \textit{in-neighbor} and \textit{out-neighbor}, which will not be considered in this paper.

\subsection{Problem Statement}

In this paper, we will study the semi-supervised node classification problem via graph representation learning. Based on the notations and terminologies defined above, we can provide studied problem formulation in this paper as follows.

\begin{definition}
(Semi-Supervised Node Classification): Given the input graph $G = (\mc{V}, \mc{E}, w, x, y)$ with a subset of partially labeled nodes $\mc{T} \subset \mc{V}$, the node classification problem studied in this paper aims at learning a mapping $f: \mc{V} \to \mc{Y}$ to project the nodes to their desired labels. To be more specific, in building the mapping $f$, we aim to to learn the representations of nodes based on both nodes' self-raw features and the diffused information from their nearby neighbors. What's more, the mapping $f$ should also be generalizable to deep architectures without suffering from the suspended animation problem.
\end{definition}
\section{Method}\label{sec:method}

To address the semi-supervised node classification problem, we will introduce the {\our} (deep diffusive neural network) model in this part. We will first talk about the {\gdu} neuron used in {\our}, and also provide necessary information analyzing its effectiveness in addressing the suspended animation problem. For the neighborhood information diffusion, we will propose an attention mechanism for information propagation in {\our}. Finally, detailed information about learning the {\our} model will be introduced at the end.

\subsection{Gated Diffusive Unit (\gdu)}

\begin{figure}
    \centering
    \begin{minipage}{.4\textwidth}
    	\includegraphics[width=\linewidth]{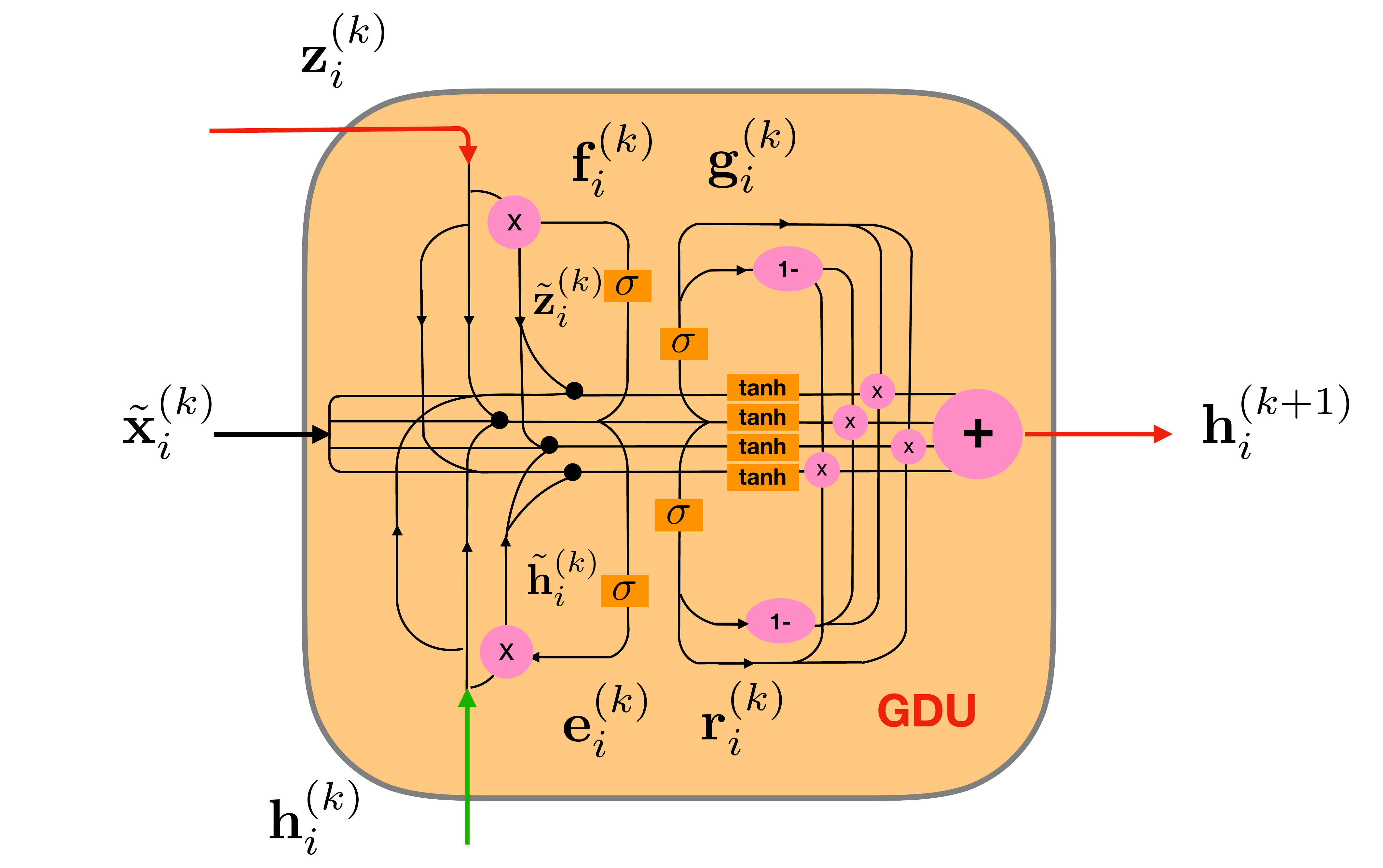}
     \end{minipage}%
        \caption{Detailed architecture of the {\gdu} neuron (for node $v_i$).}
    	\label{fig:unit}
\end{figure}

{\gdu} is a new neuron model proposed for graph representation learning specifically. Formally, given the input graph data $G$ involving the node set $\mc{V}$ and link set $\mc{E}$, according to Section~\ref{sec:formulate}, we can denote the raw feature vector and neighbor set of node $v_i$ as $\mb{x}_i$ and $\Gamma(v_i)$, respectively. The {\our} model to be build involves multiple layers, and we can denote the hidden state of node $v_i$ at the $k_{th}$ layer of {\our} as $\mb{h}_i^{(k)} \in \mathbbm{R}^{d_h}$. For the nearby neighbors of $v_i$, we can denote their hidden states as $\{\mb{h}^{(k)}_j \}_{v_j \in \Gamma(v_i)}$. {\our} introduce an attention based diffusion operator to propagate and aggregate the information among such neighbor nodes, which can be denoted as
\begin{equation}\label{equ:aggregate}
\mb{z}_i^{(k)} = \mbox{Diffusion} \left(\left\{ \mb{h}_j^{(k)} \right\}_{v_j \in \Gamma(v_i)} \right).
\end{equation}
The above operator will be defined in detail in Section~\ref{subsec:aggregation}.

As illustrated in Figure~\ref{fig:unit}, the {\gdu} (e.g., for node $v_i$) has multiple input portals. Besides the neighborhood aggregated information $\mb{z}_i$, {\gdu} will also accept $v_i$'s lower-layer representation $\mb{h}_i^{(k)}$ and graph residual term $\tilde{\mb{x}}_i$ as the other two inputs. For the aggregated input $\mb{z}_i^{(k)}$ from the neighbors, certain information in the vector can be useless for learning $v_i$'s new state. Therefore, {\gdu} defines an \textit{adjustment gate} $\mb{f}_i^{(k)}$ to update $\mb{z}_i^{(k)}$ as follows:
\begin{equation}
\hspace{-6pt} \tilde{\mb{z}}_{i}^{(k)} \hspace{-4pt}= \mb{f}_i^{(k)} \otimes \mb{z}_i^{(k)}, \mbox{and  }\mb{f}_i^{(k)} \hspace{-2pt} = \hspace{-2pt} \sigma \hspace{-2pt} \left( \mb{W}_f \left[\tilde{\mb{x}}_i \sqcup \mb{z}_i^{(k)} \sqcup \mb{h}_i^{(k)} \right] \right),
\end{equation}
Here, $\sigma(\cdot)$ is the sigmoid function parameterized by $\mb{W}_f$. 

Meanwhile, for the representation input from the lower-layer of $v_i$, i.e., $\mb{h}_i^{(k)}$, {\gdu} also introduces a similar \textit{evolving gate} $\mb{e}_i$ for adjusting the hidden state vector:
\begin{equation}
\hspace{-7pt} \tilde{\mb{h}}_i^{(k)} \hspace{-4pt} = \mb{e}_i^{(k)} \hspace{-2pt} \otimes \hspace{-1pt} \mb{h}_i^{(k)}, \mbox{and  }\mb{e}_i^{(k)} \hspace{-2pt} = \hspace{-2pt}  \sigma \hspace{-2pt} \left( \mb{W}_e \left[\tilde{\mb{x}}_i^{(k)} \hspace{-2pt} \sqcup \hspace{-2pt} \mb{z}_i^{(k)} \hspace{-2pt} \sqcup \hspace{-2pt} \mb{h}_i^{(k)} \right] \right).
\end{equation}

In addition, to involve the graph residual learning \cite{Zhang_GResNet_19} in {\our} to overcome the suspended animation problem, {\gdu} can also accept the graph residual terms as the third input, which can be denoted as
\begin{equation}
\hspace{-5pt}\tilde{\mb{x}}_i^{(k)} = \mbox{Graph-Residual}\left(\left\{\mb{x}_j \right\}_{v_j \in \mc{V}}, \left\{\mb{h}_j^{(k)} \right\}_{v_j \in \mc{V}}; G \right).
\end{equation}
Formally, $\tilde{\mb{x}}_i^{(k)}$ can represent any graph residual terms defined in \cite{Zhang_GResNet_19}. For instance, if the \textit{naive residual term} is adopted here, we will have $\tilde{\mb{x}}_i^{(k)} = \mb{h}_i^{(k)}$; whereas for the \textit{raw residual term}, we have $\tilde{\mb{x}}_i^{(k)} = {\mb{x}}_i$. For the \textit{graph-naive residual term} and \textit{graph-naive residual term}, $\tilde{\mb{x}}_i^{(k)}$ can be defined accordingly based on the graph adjacency matrix. In this paper, we will use the \textit{graph-raw residual term}, as it can achieve the best performance as shown in \cite{Zhang_GResNet_19}.

Based on these three inputs, $\tilde{\mb{x}}_i^{(k)}$, $\tilde{\mb{z}}_{i}^{(k)}$ and $\tilde{\mb{h}}_i^{(k)}$, {\gdu} can effectively integrate the useful information from them to update the hidden state of node $v_i$. Instead of simple summation, integration of such information is achieved in {\gdu} via two \textit{selection gates} $\mb{g}_i$ and $\mb{r}_i$ denoted as follows: \begingroup\makeatletter\def\f@size{8}\check@mathfonts 
\begin{equation}\label{equ:gdu}
\begin{aligned}
 \hspace{-5pt} \mb{h}_i^{(k+1)}  \hspace{-2pt} &= \mbox{\small GDU}\left( \mb{z}_i, \mb{h}_i, \tilde{\mb{x}}_i; G \right)\\
 &= \mb{g}_i^{(k)} \hspace{-2pt} \otimes \hspace{-2pt} \mb{r}_i^{(k)} \hspace{-2pt} \otimes \hspace{-2pt} \tanh \left(\mb{W}_u  \hspace{-2pt} \left[\tilde{\mb{x}}_i^{(k)} \hspace{-2pt} \sqcup \tilde{\mb{z}}_i^{(k)}  \hspace{-2pt} \sqcup \tilde{\mb{h}}_i^{(k)} \right] \right) \\
&+ (\mb{1} - \mb{g}_i^{(k)}) \hspace{-2pt} \otimes \hspace{-2pt} \mb{r}_i^{(k)} \hspace{-2pt} \otimes \hspace{-2pt} \tanh \left(\mb{W}_u  \hspace{-2pt} \left[\tilde{\mb{x}}_i^{(k)}  \hspace{-2pt} \sqcup \mb{z}_i^{(k)}  \hspace{-2pt} \sqcup \tilde{\mb{h}}_i^{(k)} \right] \right)\\
&+ \mb{g}_i^{(k)} \hspace{-2pt} \otimes \hspace{-2pt} (\mb{1} - \mb{r}_i^{(k)}) \hspace{-2pt} \otimes \hspace{-2pt} \tanh \left(\mb{W}_u  \hspace{-2pt} \left[\tilde{\mb{x}}_i^{(k)}  \hspace{-2pt} \sqcup \tilde{\mb{z}}_i^{(k)}  \hspace{-2pt} \sqcup \mb{h}_i^{(k)} \right] \right)\\
&+ (\mb{1}  \hspace{-1pt} -  \hspace{-1pt} \mb{g}_i^{(k)}) \hspace{-2pt} \otimes \hspace{-2pt} (\mb{1}  \hspace{-1pt} -  \hspace{-1pt} \mb{r}_i^{(k)})  \hspace{-2pt}  \otimes  \hspace{-2pt} \tanh  \hspace{-2pt} \left(\mb{W}_u  \hspace{-2pt} \left[\tilde{\mb{x}}_i^{(k)} \hspace{-2pt} \sqcup \mb{z}_i^{(k)}  \hspace{-2pt} \sqcup \mb{h}_i^{(k)} \right] \right),
\end{aligned}
\end{equation}
where the selection gates
\begin{equation}\label{equ:selection_gate}
\begin{cases}
\mb{g}_i^{(k)}  &= \sigma \left( \mb{W}_g  \hspace{-2pt} \left[\tilde{\mb{x}}_i^{(k)}  \sqcup \mb{z}_i^{(k)} \sqcup \mb{h}_i^{(k)}  \sqcup \tilde{\mb{z}}_i^{(k)} \sqcup \tilde{\mb{h}}_i^{(k)} \right] \right); \\
\mb{r}_i^{(k)} &= \sigma \left( \mb{W}_r  \hspace{-2pt} \left[\tilde{\mb{x}}_i^{(k)}  \sqcup \mb{z}_i^{(k)} \sqcup \mb{h}_i^{(k)} \sqcup \tilde{\mb{z}}_i^{(k)} \sqcup \tilde{\mb{h}}_i^{(k)} \right] \right).
\end{cases}
\end{equation}\endgroup
In the above equation, term $\mb{1}$ denotes a vector filled with value $1$ of the same dimensions as the \textit{selection gate} vectors $\mb{g}_i^{(k)}$ and $\mb{r}_i^{(k)}$. Operator $\tanh(\cdot)$ denotes the hyperbolic tangent activation function and $\otimes$ denotes the entry-wise product as introduced in Section~\ref{subsec:notation}.


\subsection{More Discussions on {\gdu}}

Here, we would like to provide more discussions about the {\gdu} neuron, and also introduce a simplified version of {\gdu} to lower down the learning cost of {\our}.

\subsubsection{Design of {\gdu} on Suspended Animation}

The design of the {\gdu} architecture can help {\our} solve the ``suspended animation problem'' when the model goes deep in three perspectives:

\begin{itemize}
\item \textit{Anti-Suspended Animation}: Instead of integrating the neighborhood information with the approximated graph convolutional operator via either equal importance (as {\gcn}) or with weighted summation (as {\gat}), {\gdu} introduces (a) an attentive neighborhood information diffusion and integration mechanism (to be introduced in the following subsection), and (b) a node state adjustment mechanism via four neural gates defined above. {\gdu} is not based on the approximated graph convolutional operator at all. Both the attentive diffusion mechanism and neural gates based adjustment mechanism are dynamic, which can compute the parameters automatically depending on the specific learning scenarios and can overcome the suspended animation side-effects on the model learning performance.

\item \textit{Residual Learning}: As introduced in \cite{Zhang_GResNet_19}, the graph residual terms (especially the \textit{raw residual term} and \textit{graph-raw residual term}) can help the GNN models address the ``suspended animation problem''. Both theoretic proofs and empirical experiments have demonstrated its effectiveness. The architecture of {\gdu} also effectively integrate the graph residual learning mechanism into the learning process, where the input portal $\tilde{\mb{x}}_i^{(k)}$ can accept various graph residual terms. 

\item \textit{Gradient Exploding/Vanishing}: In addition, as introduced in \cite{Hochreiter_Long_97}, for the neural network models in a deep architecture, the learning process may also suffer from the gradient exploding/vanishing problem. Similar problem has been observed for the GNN models, which is different from the ``suspended animation problem'' that we mention above. Meanwhile, similar to LSTM \cite{Hochreiter_Long_97} and GRU \cite{Chung_Empirical_14}, the neural gates introduced in {\gdu} can help ease even solve the ``gradient exploding/vanishing'' problem effectively.

\end{itemize}

Viewed in such perspectives, {\gdu} is very different from the neurons used in the existing GNN models. In the following part, we will focus on applying {\gdu} to construct the {\our} model for the graph data representation learning.

\begin{figure}
    \centering
    \begin{minipage}{.5\textwidth}
    	\includegraphics[width=\linewidth]{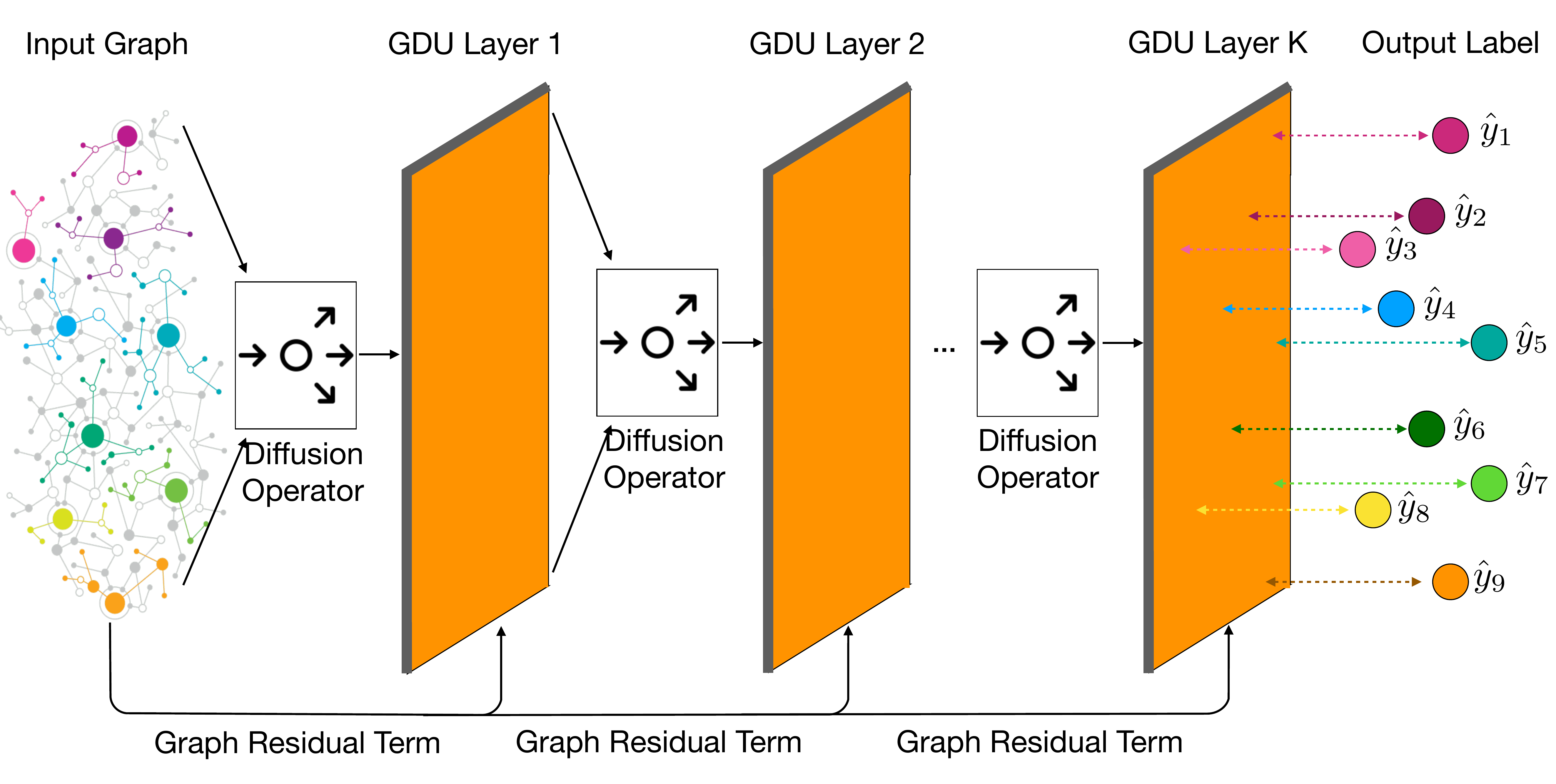}
     \end{minipage}%
        \caption{Framework architecture of the {\our} model.}
    	\label{fig:framework}
\end{figure}

\subsubsection{Simplified {\gdu}}

Meanwhile, we also observed that the {\gdu} neuron involves a complex architecture involving multiple variables, which may lead to a relatively high time cost for learning the model. To lower down the learning time cost, we also introduce a simplified version of {\gdu} in this paper here. The main difference of the simplified {\gdu} versus the original one introduced before lie in Equations~\ref{equ:gdu}-\ref{equ:selection_gate}, which can be denoted as follows:
\begin{equation}\label{equ:gdu2}
\begin{aligned}
 \hspace{-5pt} \mb{h}_i^{(k+1)}  &= \mbox{GDU}\left(\mb{z}_i, \mb{h}_i, \tilde{\mb{x}}_i; G \right)\\
 &=\tanh \Big( \mb{g}_i^{(k)} \otimes \mb{W}_u  [\tilde{\mb{z}}_i^{(k)} \sqcup \tilde{\mb{h}}_i^{(k)} ]  \\
&+ (\mb{1} - \mb{g}_i^{(k)})  \otimes  \mb{W}_u [\mb{z}_i^{(k)} \sqcup \mb{h}_i^{(k)} ] + \mb{W}'_u \tilde{\mb{x}}_i \Big),
\end{aligned}
\end{equation}
where the selection gates
\begin{equation}\label{equ:selection_gate2}
\mb{g}_i^{(k)}  = \sigma \left( \mb{W}_g \left[ \tilde{\mb{z}}_i^{(k)} \sqcup \tilde{\mb{h}}_i^{(k)} \right] \right).
\end{equation}
According to the equations, one of the selection gate $\mb{r}_i^{(k)}$ defined in Equation~\ref{equ:selection_gate} is removed, and the definition of gate $\mb{g}_i^{(k)}$ is also simplified, which is determined by the adjusted representations $\tilde{\mb{z}}_i^{(k)}$ and $\tilde{\mb{h}}_i^{(k)}$ only. Also the architecture of {\gdu} is also updated, where the node state updating equation will balance between $\mb{W}_u  [\tilde{\mb{z}}_i^{(k)} \sqcup \tilde{\mb{h}}_i^{(k)} ]$ and $\mb{W}_u [\mb{z}_i^{(k)} \sqcup \mb{h}_i^{(k)}]$ (controlled by the gate $\mb{g}_i^{(k)}$). The residual term $\mb{W}'_u \tilde{\mb{x}}_i$ is added to the representation at the end only, where $\mb{W}'_u$ is the variable for vector dimension adjustment, which can be shared across layers with the same hidden dimensions.

\subsection{Attention based Neighborhood Diffusion}\label{subsec:aggregation}

In this part, we will define the $\mbox{Diffusion}(\cdot)$ operator used in Equation~(\ref{equ:aggregate}) for node neighborhood information integration. The {\our} model defines such an operator based on an attention mechanism. Formally, given the nodes hidden states $\{\mb{h}_i^{(k)}\}_{v_i \in \mc{V}}$, {\our} defines the diffusion operator together with the nodes integrated representations as follows:
\begin{equation}
\begin{aligned}
\hspace{-5pt} \mb{Z}^{(k)} \hspace{-2pt} &= \mbox{ Diffusion} \left( \left\{\mb{h}_j^{(k)} \right\}_{v_j \in \mc{V}} \right)\\
& =\mbox{\small softmax} \left( \mb{M} \otimes \frac{ \mb{Q} \mb{K}^\top }{\sqrt{d_h}} \right) \mb{V} , 
\end{aligned}
\end{equation}
where $\mb{V} = \left[\mb{h}_1^{(k)}, \mb{h}_2^{(k)}, \cdots, \mb{h}_{|\mc{V}|}^{(k)} \right]^\top  \in \mathbbm{R}^{|\mc{V}| \times d_h}$ covers the hidden states of all the nodes in the input graph. Matrices $\mb{Q}$ and $\mb{K}$ are the identical copies of $\mb{V}$. The output matrix $\mb{Z}^{(k)} = \left[\mb{z}_1^{(k)}, \mb{z}_2^{(k)}, \cdots, \mb{z}_{|\mc{V}|}^{(k)} \right]^\top \in \mathbbm{R}^{|\mc{V}| \times d_z}$ denotes the aggregated neighbor representations of all the nodes. Term $\mb{M} \in \{0, 1\}^{|\mc{V}| \times |\mc{V}|}$ denotes a mask matrix, where entry $\mb{M}(i,j) = 1$ iff $(v_i, v_j) \in \mc{E} \lor (v_j, v_i) \in \mc{E} \lor v_i = v_j$. Matrix $\mb{M}$ can effectively help prune the influences among nodes which are not connected in the original input graph.

According to the above equation, {\our} quantifies the influence of node $v_j$ on $v_i$ in the $k_{th}$ layer based on their hidden states. If we expand the above equation, given $v_i$ and its adjacent neighbor set $\Gamma(v_i)$, we can also rewrite the learned $v_i$'s aggregated neighborhood representation vector as
\begin{equation}
\begin{aligned}
\mb{z}_i^{(k)} &= \mbox{Diffusion}\left( \left\{ \mb{h}_j^{(k)} \right \}_{v_j \in \Gamma(v_i) \cup \{v_i\} } \right)\\
&= \sum_{v_j \in \Gamma(v_i) \cup \{v_i\}} \bs{\omega}^{(k)}_{i}(j) \cdot \mb{h}_j^{(k)},
\end{aligned}
\end{equation}
where the influence weight $\bs{\omega}^{(k)}_i$ is defined as
\begin{equation}
\begin{aligned}
&\bs{\omega}^{(k)}_i = \mbox{softmax} \left( \left[e^{(k)}_{1,i}, e^{(k)}_{2,i}, \cdots, e^{(k)}_{|\mc{V}|,i} \right] \right),\\
&e^{(k)}_{j,i} = \begin{cases}
\frac{1}{\sqrt{d_h}} (\mb{h}_j^{(k)})^\top \mb{h}_i^{(k)}, & \mbox{ if } v_j \in \Gamma(v_i) \cup \{v_i\}; \\
0, & \mbox{ otherwise}.
\end{cases}
\end{aligned}
\end{equation}

%
%

\subsection{Graph Diffusive Neural Network Learning}

Based on both the {\gdu} neurons and the attentive diffusion operator, we can construct the {\our} model, whose detailed architecture is provided in Figure~\ref{fig:framework}. Based on the input graph, {\our} involves multiple layers of {\gdu} neurons, which accept inputs from the previous layer, diffusion operators and the graph residual terms. In {\our}, at each layer of the model, one {\gdu} neuron is constructed to represent the state of each node. Based on the learned representations of each node at the last layer, one fully-connected layer is adopted to project the node representations to their labels. Formally, $\forall v_i \in \mc{V}$, according to Figure~\ref{fig:framework}, its learning process of {\our} can be denoted as the following equations:\begingroup\makeatletter\def\f@size{8.0}\check@mathfonts
\begin{equation}
\hspace{-10pt}\begin{cases}
&\hspace{-8pt}{\small \mbox{/* Initialization: */}}\\
&\hspace{-8pt}\mb{x}_i = \mbox{\small Embedding}(\mb{x}_i), \mbox{ and } \mb{h}_i^{(0)} = \mbox{\small Linear}(\mb{x}_i, \mb{W}_x); \\
&\hspace{-8pt}\tilde{\mb{x}}_i = \mbox{\small Graph-Residual}\left(\left\{\mb{x}_j \right\}_{v_j \in \mc{V}}, \left\{\mb{h}_j^{(k)} \right\}_{v_j \in \mc{V}}; G \right); \\
&\hspace{-8pt}{\small \mbox{/* Iterative Updating: */}}\\
&\hspace{-12pt}\begin{cases}
&\hspace{-10pt} \mb{z}_i^{(k)} = \mbox{\small Diffusion} \left(\{\mb{h}_j^{(k)}\}_{v_j \in \Gamma(v_i) \cup \{v_i\}} \right), \forall k \in \{0, \cdots, K\}; \\
&\hspace{-10pt} \mb{h}_i^{(k+1)} \hspace{-2pt} = \mbox{\small GDU}\left( \mb{z}_i, \mb{h}_i, \tilde{\mb{x}}_i; G \right), \forall k \in \{0, \cdots, K\};\\
\end{cases}\\
&\hspace{-8pt}{\small \mbox{/* Output: */}}\\
&\hspace{-8pt}\hat{\mb{y}}_i = \mbox{\small softmax} \left(\mb{W}_{fc} \mb{h}_i^{(K+1)} \right).
\end{cases}
\end{equation}\endgroup
In the above equation, notation $\mbox{Embedding}(\cdot)$ will embed the nodes' raw input features (usually in a large dimension) into a latent representation. Depending on the raw features, different models can be used to define the $\mbox{Embedding}(\cdot)$ function. For instance, CNN can be used if $\mb{x}_i$ is an image; LSTM/RNN can be adopted if $\mb{x}_i$ is text; and fully connected layers can be used if $\mb{x}_i$ is just a long feature vector. The nodes' hidden states $\mb{h}_i^{(0)}$ is also initialized based on the embedded raw feature vectors with a linear layer parameterized by $\mb{W}_x$. The learning process involves several diffusion and {\gdu} layers, which will update nodes' representations iteratively. The final outputted latent state vectors will be projected to the nodes' labels with a fully-connected layer (also normalized by the softmax function).

Based on the set of labeled nodes, i.e., the training set $\mc{T}$, {\our} computes the introduced loss by comparing the learned label against the ground-truth, which can be denoted as the following equation: \vspace{-9pt}
\begin{equation}
\ell(\bs{\Theta}) = \sum_{v_i \in \mc{T}}  \sum_{d=1}^{d_y} - \mb{y}_i(d) \log \hat{\mb{y}}_i(d). \vspace{-5pt}
\end{equation}
By minimizing the above loss term, {\our} can be effectively learned with the error back-propagation algorithm.

\begin{table}[t]
\caption{Learning result by models with deeper architectures.}\label{tab:suspended_animation}
\centering
\small
\setlength{\tabcolsep}{4pt}
\begin{tabular}{l c c  c  c  c }
\toprule
 \multirow{2}{*}{Method (Dataset)}  & \multicolumn{5}{c}{Model Depth (Accuracy)} \\
 \addlinespace[0.05cm]
\cline{2-6}
\addlinespace[0.05cm]
& {10} & {20} & {30} & {40} & {50} \\
\addlinespace[0.05cm]
\hline
\addlinespace[0.05cm]

{\gcn} (Cora)  &0.215  &0.057  &0.057 &{0.057} &{0.057} \\

{\our} (Cora) &\textbf{0.851}  &{0.833}  &{0.834} &{0.825} &{0.825} \\
\hline
\addlinespace[0.05cm]
{\gcn} (Citeseer)&0.064  &0.064  &0.064 &{0.064} &{0.064} \\

{\our} (Citeseer) &0.650  &{0.665}  &{0.663} &{0.657} &{0.647} \\
\hline
\addlinespace[0.05cm]
{\gcn} (Pubmed) &0.366 &0.366  &0.366 &{0.366} &{0.366} \\

{\our} (Pubmed) &0.790  &{0.788}  &{0.785} &{0.787} &{0.779}\\
\bottomrule
\end{tabular}
\vspace{-10pt}
\end{table}

\section{Experiments}\label{sec:experiment}
 
To test the effectiveness of the proposed {\our} model, extensive experiments have been done on several graph benchmark datasets in this paper, including Cora, Citeseer, and Pubmed. Cora, Citeseer and Pubmed are the benchmark datasets used for semi-supervised node classification problem used in almost all the state-of-the-art GNN papers \cite{Kipf_Semi_CORR_16,Velickovic_Graph_ICLR_18,Zhang_GResNet_19}. In the experiments, for fair comparison, we will follow the identical experimental settings (e.g., train/validation/test partition) as \cite{Kipf_Semi_CORR_16}.

\noindent \textbf{Reproducibility}: Both the datasets and source code used can be accessed via link\footnote{https://github.com/jwzhanggy/DifNet}. Detailed information about the server used to run the model can be found at the footnote\footnote{GPU Server: ASUS X99-E WS motherboard, Intel Core i7 CPU 6850K@3.6GHz (6 cores), 3 Nvidia GeForce GTX 1080 Ti GPU (11 GB buffer each), 128 GB DDR4 memory and 128 GB SSD swap. For the deep models which cannot fit in the GPU memory, we run them with CPU instead.}.

\noindent \textbf{Experimental Settings}: By default, the experimental settings of the {\our} model is as follows on all these datasets: learning rate: 0.01 (0.005 on pubmed), max-epoch: 1000, dropout rate: 0.5, weight-decay: 5e{-4}, hidden size: 16, graph residual term: graph-raw.

\subsection{Model Depth vs. Suspended Animation}
 
As illustrated by Figure~\ref{fig:difnet_acc_analysis} provided in the introduction section, we have shown that {\our} model can converge very fast even with less epochs than {\gcn}. What's more, as the model depth increases, we didn't observe the suspended animation problem with {\our} at all, which can perform well even with very depth architectures, e.g., $50$ layers.

Furthermore, in Table~\ref{tab:suspended_animation}, we also provide the learning accuracy of {\our} (with different depths) on Cora, Citeseer and Pubmed, respectively. To illustrate its performance advantages, we also add the results of {\gcn} (with different depths) on these three datasets in the table as well. The results demonstrate that {\our} can also work well in handling the suspended animation problem for the other datasets from different sources as well.

\begin{table}[t]
\caption{Learning result accuracy of node classification methods. In the table, `-' denotes the results of the methods on these datasets are not reported in the existing works. We have run {\our} with depth values from $\{2, 3, \cdots, 10, 20, \cdots, 50\}$. Results reported of {\our} in this paper denotes the best one among all these depths.}\label{tab:complete_performance_comparison}
\centering
\small
\setlength{\tabcolsep}{4pt}
\begin{tabular}{l c c c c }
\toprule
 \multirow{2}{*}{Methods}  & \multicolumn{3}{c}{Datasets (Accuracy)} \\
\cline{2-4}
\addlinespace[0.05cm]
& \textbf{Cora} & \textbf{Citeseer} & \textbf{Pubmed} \\
\hline
\addlinespace[0.05cm]

{LP } &0.680 &0.453 &0.630  \\
{ICA (\cite{LG03})} &0.751  &0.691  &0.739   \\
{ManiReg (\cite{BNS06})} &0.595  &0.601  &0.707   \\
{SemiEmb (\cite{WRC08})} &0.590  &0.596  &0.711  \\
\hline
\addlinespace[0.05cm]
{DeepWalk (\cite{PAS14})} &0.672  &0.432  &0.653   \\
{Planetoid (\cite{YCS16})} &0.757  &0.647  &0.772  \\
{MoNet (\cite{MBMRSB16})} &0.817  &-  &0.788  \\
\hline
\addlinespace[0.05cm]
{{\gcn} (\cite{Kipf_Semi_CORR_16})} &0.815  &0.703  &0.790   \\
{{\gat} (\cite{Velickovic_Graph_ICLR_18})} &0.830  &0.725  &0.790  \\
{{\loopy} (\cite{loopynet})} &0.826  &0.716  &0.792  \\
{SF-GCN (\cite{DBLP:journals/corr/abs-1907-02586})} &0.833  &\textbf{0.734}  &\textbf{0.793}   \\
\hline
\addlinespace[0.05cm]

{\our} (original {\gdu}) &\textbf{0.851}  &\textbf{0.727}  &{0.783}  \\
{\our} (simplified {\gdu}) &\textbf{0.834}  &{0.715}  &\textbf{0.795}  \\

\bottomrule
\toprule
\addlinespace[0.1cm]
& \multicolumn{3}{c}{Time Cost (seconds)} \\
\cline{2-4}
\addlinespace[0.05cm]
{\our} (original {\gdu}) &47.55  &34.40  &215.40 \\
{\our} (simplified {\gdu}) &23.34  &24.96  &113.77 \\
\bottomrule
\end{tabular} \vspace{-10pt}
\end{table}

\vspace{-5pt}
\subsection{Comparison with State-of-the-Art}
 
Besides the results showing that {\our} can perform well in addressing the suspended animation problem, to make the experimental results more complete, we also compare {\our} with both classic and state-of-the-art baseline models, whose results are provided in Table~\ref{tab:complete_performance_comparison}. According to the results, compared against these baseline methods, {\our} can outperform them with great advantages. Also for {\our} with the simplified {\gdu}, it can achieve comparable performance as {\our} with the original {\gdu} with the same learning epochs. However, the learning time cost with simplified {\gdu} is greatly reduced according to the experiment results.

\vspace{-7pt}
\section{Conclusion}\label{sec:conclusion}
\vspace{-4pt}

In this paper, we focus on studying the graph representation learning problem and apply the learned graph representations for node semi-supervised classification. To address the common ``suspended animation problem'' with the existing graph neural networks, we introduce a new graph neural network model, namely {\our}, in this paper. {\our} is constructed based on both the {\gdu} neurons and the attentive diffusion operator. To handle the suspended animation problem, {\gdu} includes both neural gate learning and graph residual learning into its architecture, which can also handle the conventional gradient vanishing/exploding problem as well. Extensive experiments have been done on several benchmark graph datasets for node classification, whose results also demonstrate the effectiveness of both {\our} and {\gdu} in deep graph representation learning.
\newpage
{\small
\bibliographystyle{named}
\bibliography{reference}

\begin{thebibliography}{}

\bibitem[\protect\citeauthoryear{Atwood and Towsley}{2015}]{AT16}
James Atwood and Don Towsley.
\newblock Search-convolutional neural networks.
\newblock {\em CoRR}, abs/1511.02136, 2015.

\bibitem[\protect\citeauthoryear{Belkin \bgroup \em et al.\egroup
  }{2006}]{BNS06}
Mikhail Belkin, Partha Niyogi, and Vikas Sindhwani.
\newblock Manifold regularization: A geometric framework for learning from
  labeled and unlabeled examples.
\newblock {\em J. Mach. Learn. Res.}, 7:2399--2434, December 2006.

\bibitem[\protect\citeauthoryear{Chung \bgroup \em et al.\egroup
  }{2014}]{Chung_Empirical_14}
Junyoung Chung, {\c{C}}aglar G{\"{u}}l{\c{c}}ehre, KyungHyun Cho, and Yoshua
  Bengio.
\newblock Empirical evaluation of gated recurrent neural networks on sequence
  modeling.
\newblock {\em CoRR}, abs/1412.3555, 2014.

\bibitem[\protect\citeauthoryear{Defferrard \bgroup \em et al.\egroup
  }{2016}]{Defferrard_Convolutional_16}
Micha{\"{e}}l Defferrard, Xavier Bresson, and Pierre Vandergheynst.
\newblock Convolutional neural networks on graphs with fast localized spectral
  filtering.
\newblock {\em CoRR}, abs/1606.09375, 2016.

\bibitem[\protect\citeauthoryear{Gao and Glowacka}{2016}]{Gao2016DeepGR}
Yuan Gao and Dorota Glowacka.
\newblock Deep gate recurrent neural network.
\newblock In {\em ACML}, 2016.

\bibitem[\protect\citeauthoryear{Gao \bgroup \em et al.\egroup
  }{2019}]{Gao_GraphNAS_19}
Yang Gao, Hong Yang, Peng Zhang, Chuan Zhou, and Yue Hu.
\newblock Graphnas: Graph neural architecture search with reinforcement
  learning.
\newblock {\em CoRR}, abs/1904.09981, 2019.

\bibitem[\protect\citeauthoryear{G{\"u}rel \bgroup \em et al.\egroup
  }{2019}]{Merve_An_19}
Nezihe~Merve G{\"u}rel, Hansheng Ren, Yujing Wang, Hui Xue, Yaming Yang, and
  Ce~Zhang.
\newblock An anatomy of graph neural networks going deep via the lens of mutual
  information: Exponential decay vs. full preservation.
\newblock {\em ArXiv}, abs/1910.04499, 2019.

\bibitem[\protect\citeauthoryear{Hammond \bgroup \em et al.\egroup
  }{2011}]{Hammond_Wavelets_11}
David~K. Hammond, Pierre Vandergheynst, and Remi Gribonval.
\newblock Wavelets on graphs via spectral graph theory.
\newblock {\em Applied and Computational Harmonic Analysis}, 30(2):129--150,
  Mar 2011.

\bibitem[\protect\citeauthoryear{He \bgroup \em et al.\egroup
  }{2015}]{He_Deep_15}
Kaiming He, Xiangyu Zhang, Shaoqing Ren, and Jian Sun.
\newblock Deep residual learning for image recognition.
\newblock {\em CoRR}, abs/1512.03385, 2015.

\bibitem[\protect\citeauthoryear{Hochreiter and
  Schmidhuber}{1997}]{Hochreiter_Long_97}
Sepp Hochreiter and J{\"u}rgen Schmidhuber.
\newblock Long short-term memory.
\newblock {\em Neural computation}, 9(8):1735--1780, 1997.

\bibitem[\protect\citeauthoryear{Huang and Carley}{2019}]{Huang_Inductive_19}
Binxuan Huang and Kathleen~M. Carley.
\newblock Inductive graph representation learning with recurrent graph neural
  networks.
\newblock {\em CoRR}, abs/1904.08035, 2019.

\bibitem[\protect\citeauthoryear{Kipf and Welling}{2016}]{Kipf_Semi_CORR_16}
Thomas~N. Kipf and Max Welling.
\newblock Semi-supervised classification with graph convolutional networks.
\newblock {\em CoRR}, abs/1609.02907, 2016.

\bibitem[\protect\citeauthoryear{Klicpera \bgroup \em et al.\egroup
  }{2018}]{Klicpera_Personalized_18}
Johannes Klicpera, Aleksandar Bojchevski, and Stephan G{\"{u}}nnemann.
\newblock Personalized embedding propagation: Combining neural networks on
  graphs with personalized pagerank.
\newblock {\em CoRR}, abs/1810.05997, 2018.

\bibitem[\protect\citeauthoryear{Li \bgroup \em et al.\egroup
  }{2018a}]{Li_Deeper_18}
Qimai Li, Zhichao Han, and Xiao{-}Ming Wu.
\newblock Deeper insights into graph convolutional networks for semi-supervised
  learning.
\newblock {\em CoRR}, abs/1801.07606, 2018.

\bibitem[\protect\citeauthoryear{Li \bgroup \em et al.\egroup
  }{2018b}]{Li_Combinatorial_18}
Zhuwen Li, Qifeng Chen, and Vladlen Koltun.
\newblock Combinatorial optimization with graph convolutional networks and
  guided tree search.
\newblock {\em CoRR}, abs/1810.10659, 2018.

\bibitem[\protect\citeauthoryear{Li \bgroup \em et al.\egroup
  }{2019}]{Li_Can_19}
Guohao Li, Matthias M{\"{u}}ller, Ali~K. Thabet, and Bernard Ghanem.
\newblock Can gcns go as deep as cnns?
\newblock {\em CoRR}, abs/1904.03751, 2019.

\bibitem[\protect\citeauthoryear{Lin \bgroup \em et al.\egroup
  }{2019}]{DBLP:journals/corr/abs-1907-02586}
Guangfeng Lin, Jing Wang, Kaiyang Liao, Fan Zhao, and Wanjun Chen.
\newblock Structure fusion based on graph convolutional networks for
  semi-supervised classification.
\newblock {\em CoRR}, abs/1907.02586, 2019.

\bibitem[\protect\citeauthoryear{Lu and Getoor}{2003}]{LG03}
Qing Lu and Lise Getoor.
\newblock Link-based classification.
\newblock In {\em Proceedings of the Twentieth International Conference on
  International Conference on Machine Learning}, ICML'03, pages 496--503. AAAI
  Press, 2003.

\bibitem[\protect\citeauthoryear{Masci \bgroup \em et al.\egroup
  }{2015}]{MBBV15}
Jonathan Masci, Davide Boscaini, Michael~M. Bronstein, and Pierre
  Vandergheynst.
\newblock Shapenet: Convolutional neural networks on non-euclidean manifolds.
\newblock {\em CoRR}, abs/1501.06297, 2015.

\bibitem[\protect\citeauthoryear{Monti \bgroup \em et al.\egroup
  }{2016}]{MBMRSB16}
Federico Monti, Davide Boscaini, Jonathan Masci, Emanuele Rodol{\`{a}}, Jan
  Svoboda, and Michael~M. Bronstein.
\newblock Geometric deep learning on graphs and manifolds using mixture model
  cnns.
\newblock {\em CoRR}, abs/1611.08402, 2016.

\bibitem[\protect\citeauthoryear{Niepert \bgroup \em et al.\egroup
  }{2016}]{NAK16}
Mathias Niepert, Mohamed Ahmed, and Konstantin Kutzkov.
\newblock Learning convolutional neural networks for graphs.
\newblock {\em CoRR}, abs/1605.05273, 2016.

\bibitem[\protect\citeauthoryear{Perozzi \bgroup \em et al.\egroup
  }{2014}]{PAS14}
Bryan Perozzi, Rami Al{-}Rfou, and Steven Skiena.
\newblock Deepwalk: Online learning of social representations.
\newblock {\em CoRR}, abs/1403.6652, 2014.

\bibitem[\protect\citeauthoryear{{Scarselli} \bgroup \em et al.\egroup
  }{2009}]{SGTHM09}
Franco {Scarselli}, Marco {Gori}, Ah~Chung {Tsoi}, Markus {Hagenbuchner}, and
  Gabriele {Monfardini}.
\newblock The graph neural network model.
\newblock {\em IEEE Transactions on Neural Networks}, 20(1):61--80, Jan 2009.

\bibitem[\protect\citeauthoryear{Srivastava \bgroup \em et al.\egroup
  }{2015}]{Kumar_Highway_15}
Rupesh~Kumar Srivastava, Klaus Greff, and J{\"{u}}rgen Schmidhuber.
\newblock Highway networks.
\newblock {\em CoRR}, abs/1505.00387, 2015.

\bibitem[\protect\citeauthoryear{Sun \bgroup \em et al.\egroup
  }{2019}]{Sun_AdaGCNAG_19}
Ke~Sun, Zhouchen Lin, and Zhanxing Zhu.
\newblock Adagcn: Adaboosting graph convolutional networks into deep models.
\newblock {\em ArXiv}, abs/1908.05081, 2019.

\bibitem[\protect\citeauthoryear{Veli{\v{c}}kovi{\'{c}} \bgroup \em et
  al.\egroup }{2018}]{Velickovic_Graph_ICLR_18}
Petar Veli{\v{c}}kovi{\'{c}}, Guillem Cucurull, Arantxa Casanova, Adriana
  Romero, Pietro Li{\`{o}}, and Yoshua Bengio.
\newblock {Graph Attention Networks}.
\newblock {\em International Conference on Learning Representations}, 2018.

\bibitem[\protect\citeauthoryear{Weston \bgroup \em et al.\egroup
  }{2008}]{WRC08}
Jason Weston, Fr\'{e}d\'{e}ric Ratle, and Ronan Collobert.
\newblock Deep learning via semi-supervised embedding.
\newblock In {\em Proceedings of the 25th International Conference on Machine
  Learning}, ICML'08, pages 1168--1175, New York, NY, USA, 2008. Association
  for Computing Machinery.

\bibitem[\protect\citeauthoryear{Wu \bgroup \em et al.\egroup }{2019}]{Wu_A_19}
Zonghan Wu, Shirui Pan, Fengwen Chen, Guodong Long, Chengqi Zhang, and
  Philip~S. Yu.
\newblock A comprehensive survey on graph neural networks.
\newblock 2019.

\bibitem[\protect\citeauthoryear{Yang \bgroup \em et al.\egroup }{2016}]{YCS16}
Zhilin Yang, William~W. Cohen, and Ruslan Salakhutdinov.
\newblock Revisiting semi-supervised learning with graph embeddings.
\newblock {\em CoRR}, abs/1603.08861, 2016.

\bibitem[\protect\citeauthoryear{Zhang and Meng}{2019}]{Zhang_GResNet_19}
Jiawei Zhang and Lin Meng.
\newblock Gresnet: Graph residual network for reviving deep gnns from suspended
  animation.
\newblock {\em CoRR}, abs/1909.05729, 2019.

\bibitem[\protect\citeauthoryear{Zhang}{2018}]{loopynet}
Jiawei Zhang.
\newblock Deep loopy neural network model for graph structured data
  representation learning.
\newblock {\em CoRR}, abs/1805.07504, 2018.

\bibitem[\protect\citeauthoryear{Zhang}{2019}]{Zhang_Graph_19}
Jiawei Zhang.
\newblock Graph neural networks for small graph and giant network
  representation learning: An overview.
\newblock {\em CoRR}, abs/1908.00187, 2019.

\bibitem[\protect\citeauthoryear{Zhou \bgroup \em et al.\egroup
  }{2018}]{ZCZYLS18}
Jie Zhou, Ganqu Cui, Zhengyan Zhang, Cheng Yang, Zhiyuan Liu, and Maosong Sun.
\newblock Graph neural networks: {A} review of methods and applications.
\newblock {\em CoRR}, abs/1812.08434, 2018.

\end{thebibliography}
}


\end{document}